\documentclass[11pt,a4paper]{article}
\usepackage[hyperref]{ranlp2021}
\usepackage{times}
\usepackage{latexsym}
\usepackage{multirow}
\usepackage{multicol}
\usepackage{tabularx}
\usepackage{graphicx}
\usepackage{booktabs}
\usepackage{mathtools}

\usepackage{enumitem}
\usepackage{booktabs}
\usepackage{multirow}
\usepackage{soul}

\usepackage{subcaption}
\usepackage{epstopdf}
\usepackage{xstring}

\usepackage{balance}
\usepackage{pbox}
\usepackage{multirow}
\usepackage{enumitem}

\usepackage{multicol}

\usepackage{tabularx}
\usepackage{tcolorbox}
\usepackage{color, colortbl}

\usepackage{comment}
\newcommand{\deltext}[1]{} 

\usepackage{microtype}

\aclfinalcopy 
\def\aclpaperid{257} 

\title{A Second Pandemic?\\
Analysis of Fake News About COVID-19 Vaccines in Qatar}

\author{
Preslav Nakov, \textsuperscript{\rm 1} 
Firoj Alam,\textsuperscript{\rm 1} 
Shaden Shaar,\textsuperscript{\rm 1}
\\{\bf Giovanni {Da San Martino}\textsuperscript{\rm 2} and Yifan Zhang\textsuperscript{\rm 1}} \\
\textsuperscript{\rm 1} Qatar Computing Research Institute, HBKU, Qatar\\
\textsuperscript{\rm 2} University of Padova, Italy\\
\texttt{\{pnakov, fialam, sshaar, yzhang\}@hbku.edu.qa}, \\\texttt{dasan@math.unipd.it}
\\}

\date{}

\begin{document}
\maketitle
\begin{abstract}
While COVID-19 vaccines are finally becoming widely available, a second pandemic that revolves around the circulation of anti-vaxxer ``fake news'' may hinder efforts to recover from the first one. With this in mind, we performed an extensive analysis of Arabic and English tweets about COVID-19 vaccines, with focus on messages originating from Qatar. We found that Arabic tweets contain a lot of false information and rumors, while English tweets are mostly factual. However, English tweets are much more propagandistic than Arabic ones. In terms of propaganda techniques, about half of the Arabic tweets express doubt, and 1/5 use loaded language, while English tweets are abundant in loaded language, exaggeration, fear, name-calling, doubt, and flag-waving.  Finally, in terms of framing, Arabic tweets adopt a health and safety perspective, while in English economic concerns dominate. 
\end{abstract}

\section{Introduction}
\label{sec:introduction}

During the COVID-19 pandemic, social media have become one of the main communication channels for information dissemination and consumption, and many people rely on them as their primary source of news \citep{perrin2015social}, attracted by the broader choice of information sources. Unfortunately, over time, social media have also become one of the main channels to spread disinformation. To tackle this issue, a number of (mostly manual) fact-checking initiatives have been launched, and there are over 200 fact-checking organizations currently active worldwide.\footnote{\url{http://tiny.cc/zd1fnz}} However, these efforts are insufficient, given the scale of disinformation, which, in the time of COVID-19, has grown into the \emph{First Global Infodemic} (according to the World Health Organization).

\begin{figure}[h!]
\centering
\includegraphics[width=0.51\textwidth]{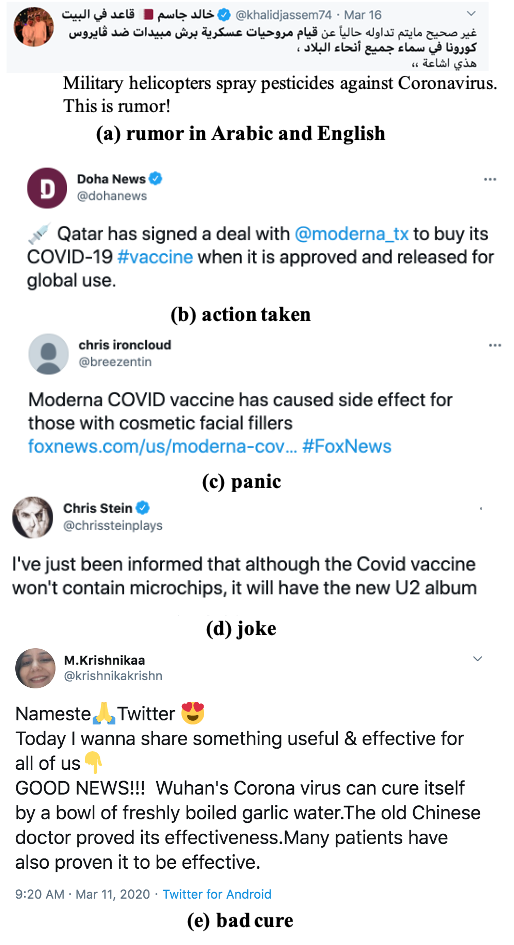}
\caption{Tweets about COVID-19 and vaccines.}
\label{fig:tweet_examples}
\end{figure}

Figure \ref{fig:tweet_examples} shows examples of how online users discuss COVID-19 and vaccines. We can see that the problem goes beyond factuality: there are tweets spreading rumors, discussing action taken, instilling panic, making jokes, and promoting bad cure.

For the tweets in Figure~\ref{fig:tweet_examples}, we might want to know whether they are factual, harmful, calling for action, etc. (see Section \ref{ssec:covid_disinfo_analysis}). It is also important to understand whether the content of the tweet is propagandistic (Section~\ref{ssec:propaganda_analysis}), what propaganda techniques are used, as well as the way the issue is framed (Section~\ref{ssec:media_framing_analysis}). Doing this in a timely manner is crucial to help organizations channel their efforts, and to counter the spread of disinformation, which may cause panic, mistrust, and other problems. 

With this in mind, we performed an extensive analysis of Arabic and English tweets about COVID-19 and vaccines, with focus on messages originating from Qatar. Our analysis focuses on (\emph{i})~COVID-19 disinformation, 
(\emph{ii})~propaganda and its techniques, and (\emph{iii})~framing.

Our contributions can be summarized as follows:
\begin{itemize}
	\item We build and release a dataset of tweets related to COVID-19 and vaccines in Arabic and English.\footnote{http://gitlab.com/sshaar/a-second-pandemic.-analysis-of-fake-news-about-covid-19-vaccines-in-qatar}
	\item We analyze the tweets from various perspectives (factuality, harmfulness, propaganda, and framing), and we discuss some interesting observations from our analysis.
\end{itemize}

\section{Related Work}
\label{sec:related_work}

Below, we discuss relevant research directions.

\subsection{Factuality}

Work on fighting disinformation and misinformation online has focused on fact-checking and fake news detection \cite{Li:2016:STD:2897350.2897352,hardalov2016,Shu:2017:FND:3137597.3137600,karadzhov2017fully,Lazer1094,aaai2018:factchecking,Vosoughi1146,vo2018rise,atanasova2019automatic,source:multitask:naacl:2019,emnlp2019:fauxtography,baly2020written,CIKM2020:FANG,shaar-etal-2020-known}. Research was further enabled by the emergence of datasets \cite{wang:2017:short,augenstein-etal-2019-multifc}, often released as part of evaluation campaigns \cite{derczynski-etal:2017:semeval,clef2018checkthat:overall,NLP4IF2019:propaganda:task,checkthat:ecir2019short,gorrell2019semeval,mihaylova-etal-2019-semeval,CheckThat:ECIR2020,ECIR2021:CLEF,CheckThat:ECIR2021,shaar-etal-2021-findings}.
As automated systems have credibility issues~\cite{fullfact:coof}, another research direction has emerged: building tools to facilitate human fact-checkers \cite{Survey:2021:AI:Fact-Checkers}.

\subsection{Check-Worthiness Estimation}

Given the volume of claims appearing in social media posts or in political statements, a problem that is crucial for fact-checkers is to identify which claims should be prioritized for fact-checking. 
The ClaimBuster system~\cite{Hassan:15} was a pioneering work  in that direction. It categorized a political statement as \textit{non-factual}, \textit{unimportant factual}, or \textit{check-worthy factual}.  \newcite{gencheva-EtAl:2017:RANLP} also focused on the 2016 US Presidential debates, for which they obtained binary (\emph{check-worthy} vs. \emph{non-check-worthy}) labels based on the fact-checking decisions of nine fact-checking organizations. An extension of this work was the ClaimRank system, which supports both English and Arabic~\cite{NAACL2018:claimrank}. 
Note that political debates and speeches require modeling the context of the target sentence to classify. Indeed, context was a major focus for most research in the debates domain \cite{gencheva-EtAl:2017:RANLP,Patwari:17,RANLP2019:checkworthiness:multitask,claim:retrieval:context:2021}. For example, \newcite{RANLP2019:checkworthiness:multitask} modeled context in a multi-task learning neural network that predicts whether a sentence would be selected for fact-checking by each fact-checking organization (from a set of nine such organizations). 

There has also been research on detecting check-worthy claims \emph{in social media} (as opposed to the above research, which targeted political debates and speeches), featuring tweets about COVID-19 or general topics in Arabic and English \cite{clef-checkthat-ar:2020,clef-checkthat-en:2020,clef-checkthat:2021:task1}. 

More directly related to our work here is the work of \newcite{alam2020fighting} and \newcite{alam2020call2arms}, who developed a multi-question annotation schema of tweets about COVID-19, organized around seven questions that aim to model the perspective of journalists, fact-checkers, social media platforms, policymakers, and the society. In our experiments, we use their schema and data to train classifiers for part of our analysis.

\subsection{Propaganda}

Propaganda is a communication tool that is deliberately designed to influence the opinions and the actions of other people in order to achieve a predetermined objective. 
When automatic means are being used to spread such influencing messages on social media platforms, this is referred to as \emph{computational propaganda} \cite{woolley2018computational}. 

Research on propaganda detection has focused on textual content \cite{BARRONCEDENO20191849,rashkin-EtAl:2017:EMNLP2017,EMNLP19DaSanMartino,da2020survey}. Suitable datasets were made available by \newcite{rashkin-EtAl:2017:EMNLP2017} and \newcite{BARRONCEDENO20191849}, where the documents (news articles) were annotated using distant supervision, according to the reputation of their source, as judged by journalists. \newcite{rashkin-EtAl:2017:EMNLP2017} focused on analyzing the language of propaganda (vs. trusted, satire, and hoaxes) based on LIWC lexicons, while \newcite{BARRONCEDENO20191849} studied a variety of stylistic features. 

\citet{Habernal.et.al.2017.EMNLP,Habernal2018b} developed a corpus annotated with five fallacies, including \textit{ad hominem}, \textit{red herring}, and \textit{irrelevant authority}. Fine-grained propaganda analysis was done by \citet{EMNLP19DaSanMartino}, who developed a corpus of news articles annotated with 18 propaganda techniques. Subsequently, the Prta system was released \cite{da2020prta}, and improved models were proposed, focusing on interpretability \cite{RANLP2021:propaganda:interpretable} or addressing the limitations of transformers \cite{ECML2021:end:of:history}. Finally, multimodal content was explored in memes using 22 propaganda techniques \cite{dimitrov2021detecting,SemEval2021-6-Dimitrov}.

\subsection{Framing} 

Framing refers to representing different salient aspects and perspectives for the purpose of conveying the latent meaning about an issue \cite{entman1993framing}. 
Recent work on automatically identifying media frames includes developing coding schemes and semi-automated methods~\cite{boydstun2013identifying}, 
datasets such as the Media Frames Corpus~\cite{card-etal-2015-media}, and systems to automatically detect media frames \cite{liu2019detecting,zhang-etal-2019-tanbih}, 
large-scale automatic analysis of news articles~\cite{kwak2020systematic}, and semi-supervised approaches~\cite{cheeks2020discovering}.

\subsection{Fighting the COVID-19 Infodemic}

Related work on fighting the COVID-19 infodemic includes developing multi-question annotation schemes of tweets about COVID-19~\cite{alam2020fighting,alam2020call2arms}, studying credibility \cite{cinelli2020covid19,pulido2020covid,zhou2020repository}, racial prejudices and fear \cite{Medford2020.04.03.20052936,vidgen2020detecting},
 situational information, e.g.,~caution and advice \citep{li2020characterizing}, as well as detecting mentions and stance with respect to known misconceptions \cite{hossain-etal-2020-covidlies}.

\section{Dataset}
\label{sec:dataset}

We collected Arabic tweets from February 2020 till March 2021.
For the English tweets, we had two separate time periods (before and after COVID-19 vaccines became available): (\emph{i})~from February till August 2020 (644 tweets), and (\emph{ii})~from November 2020 till January 2021 (1,945 tweets).
We used the following keywords to collect the tweets:

\begin{figure}[h]
\centering
\includegraphics[width=0.40\textwidth]{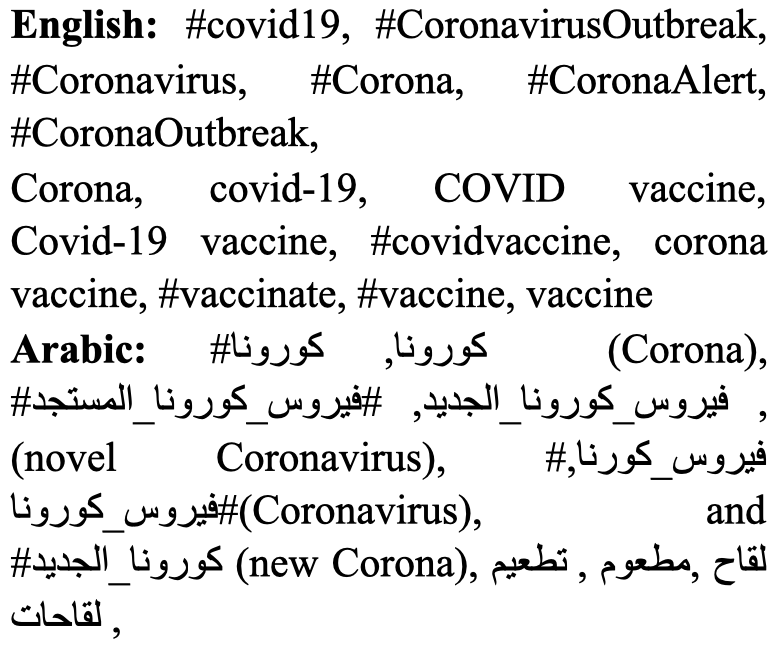}
\end{figure}

We collected original tweets (no retweets or replies), we removed the duplicates using the similarity-based approach in \newcite{alam2020standardizing}, and we filtered out tweets with less than five words. Finally, we kept the most frequently liked and retweeted tweets for annotation. Our final corpus consists of 606 Arabic and 2,589 English tweets. 

\section{Method}
\label{sec:methods}

\begin{figure}[tbh]
\centering
\includegraphics[width=0.5\textwidth]{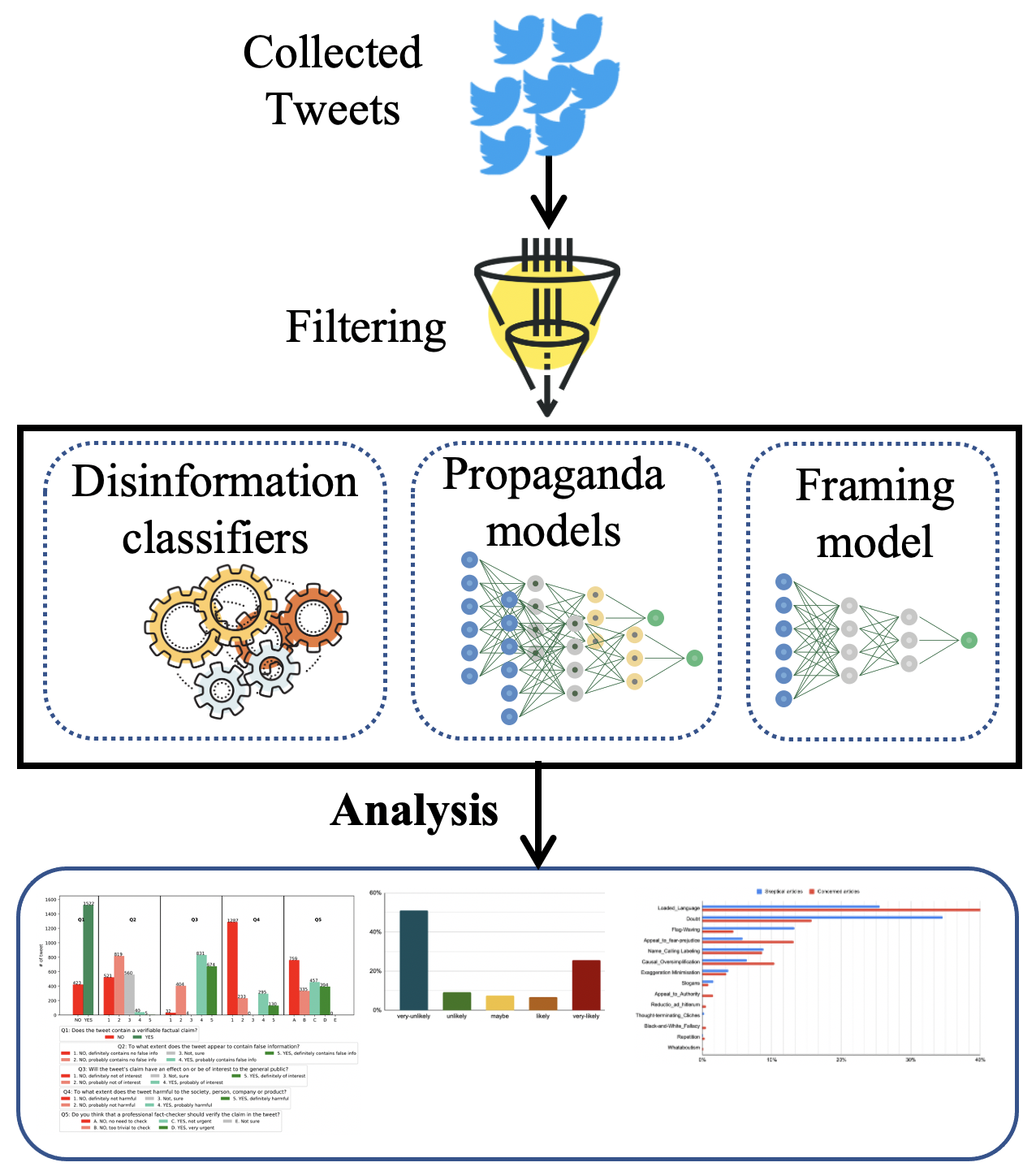}
\caption{The architecture of our system for tweet analysis. The arrows show the information flow.}
\label{fig:system_arch}
\end{figure}

Figure \ref{fig:system_arch} shows the architecture of our system. Below, we discuss each analysis step in detail.

\subsection{Disinformation Analysis}
\label{ssec:covid_disinfo_analysis}

For disinformation analysis, we used the dataset from \cite{alam2020fighting,alam2020call2arms}, which is organized around seven questions: asking whether the tweet (Q1)~contains a verifiable factual claim, (Q2)~is likely to contain false information, (Q3)~is of interest to the general public, (Q4)~is potentially harmful to a person, a company, a product, or the society, (Q5)~requires verification by a fact-checker, (Q6)~poses harm to society, or (Q7)~requires the attention of policy makers. The dataset consist of 504 English and 218 Arabic tweets, and we used it to train an SVM classifier, whose hyper-parameters we optimized using 10-fold cross-validation. Table~\ref{tab:disinfo_classification_results} shows the performance of the classifier for English and Arabic for all questions. Note the multiclass nature of the tasks and the skewed class distribution for Q2 to Q6~\cite{alam2020call2arms}.

\begin{table}[]
\centering
\scalebox{0.90}{
\begin{tabular}{@{}lrrrr@{}}
\toprule
\multicolumn{1}{c}{Q (\# Cl.)} & \multicolumn{2}{c}{\textbf{English}} & \multicolumn{2}{c}{\textbf{Arabic}} \\ \midrule
\multicolumn{1}{c}{\textbf{}} & \multicolumn{1}{c}{\textbf{Acc}} & \multicolumn{1}{c}{\textbf{W-F1}} & \multicolumn{1}{c}{\textbf{Acc}} & \multicolumn{1}{c}{\textbf{W-F1}} \\ \midrule
Q1 (2) & 64.5 & 64.8 & 72.5 & 72.9 \\
Q2 (5) & 40.0 & 41.1 & 44.3 & 43.3 \\
Q3 (5) & 42.0 & 41.7 & 51.4 & 49.1 \\
Q4 (5) & 41.6 & 41.5 & 57.1 & 56.4 \\
Q5 (5) & 37.0 & 37.6 & 57.1 & 57.4 \\
Q6 (8) & 50.0 & 50.4 & 69.7 & 68.6 \\
Q7 (10) & 60.7 & 58.6 & 68.8 & 69.1 \\ \bottomrule
\end{tabular}
}
\caption{Performance of our models for disinformation analysis. Here, \emph{\# Cl.} shows the number of classes for the corresponding question.}
\label{tab:disinfo_classification_results}
\end{table}

\subsection{Propaganda Analysis}
\label{ssec:propaganda_analysis}

For the propaganda analysis, we used two systems: Proppy \cite{aaai2019:proppy} and Prta \cite{da2020prta}. 

\textbf{Proppy} uses a maximum entropy classifier trained on 51k articles, represented with various style-related features, such as character $n$-grams and a number of vocabulary richness and readability measures. The performance of the model is 82.89 in terms of F1 score, as evaluated on a separate test set of 10k articles. It outputs the following propaganda labels based on the output score $p \in [0,1]$: \emph{very unlikely} ($0.0 \leq p < 0.2$), \emph{unlikely} ($0.2 \leq p < 0.4$), \emph{somehow} ($0.4 \leq p < 0.6$), \emph{likely} ($0.6 \leq p < 0.8$), and \emph{very likely} ($0.8 \leq p \leq 1.0$).

The \textbf{Prta system} offers a fragment-level and a sentence-level classifiers. They were trained on a corpus of 350K tokens. The performance of the sentence-level classifier is 60.71 in terms of F1 score. The fragment-level classifier identifies the text fragments and the propaganda techniques that occur in them. They consider the following 18 techniques: (\emph{i})~Loaded language, (\emph{ii})~Name calling or labeling, (\emph{iii})~Repetition, (\emph{iv})~Exaggeration or minimization, (\emph{v})~Doubt, (\emph{vi})~Appeal to fear/prejudice, (\emph{vii})~Flag-waving, (\emph{viii})~Causal oversimplification, (\emph{ix})~Slogans, (\emph{x})~Appeal to authority, (\emph{xi})~Black-and-white fallacy, dictatorship, (\emph{xii})~Thought-terminating clich\'e, (\emph{xiii})~Whataboutism, (\emph{xiv})~Reductio ad Hitlerum, (\emph{xv})~Red herring, (\emph{xvi})~Bandwagon, (\emph{xvii})~Obfuscation, intentional vagueness, confusion, and (\emph{xviii})~Straw man.

Note that both Proppy and Prta were developed for English. Thus, for the classification of Arabic content, we first translated it to English using the Google translation API, and then we ran the tools.

\begin{figure*}[tbh!]
\centering
    \begin{subfigure}[b]{0.75\textwidth}
        \includegraphics[width=\textwidth]{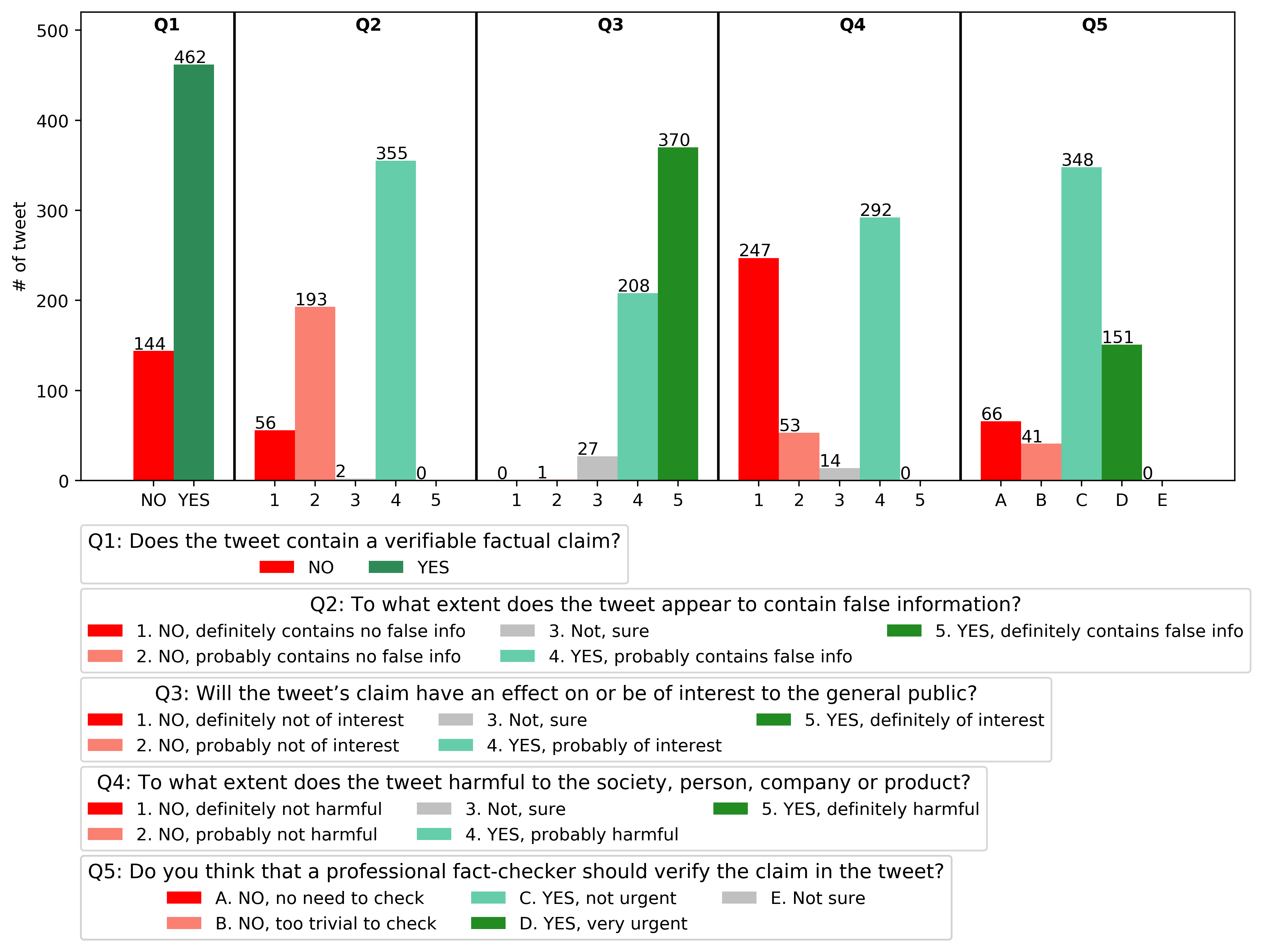}
        \caption{Questions Q1-Q5.}
        \label{fig:data_dist_group1_arabic}
    \end{subfigure}
    \hfill
    \begin{subfigure}[b]{0.75\textwidth}    
        \includegraphics[width=\textwidth]{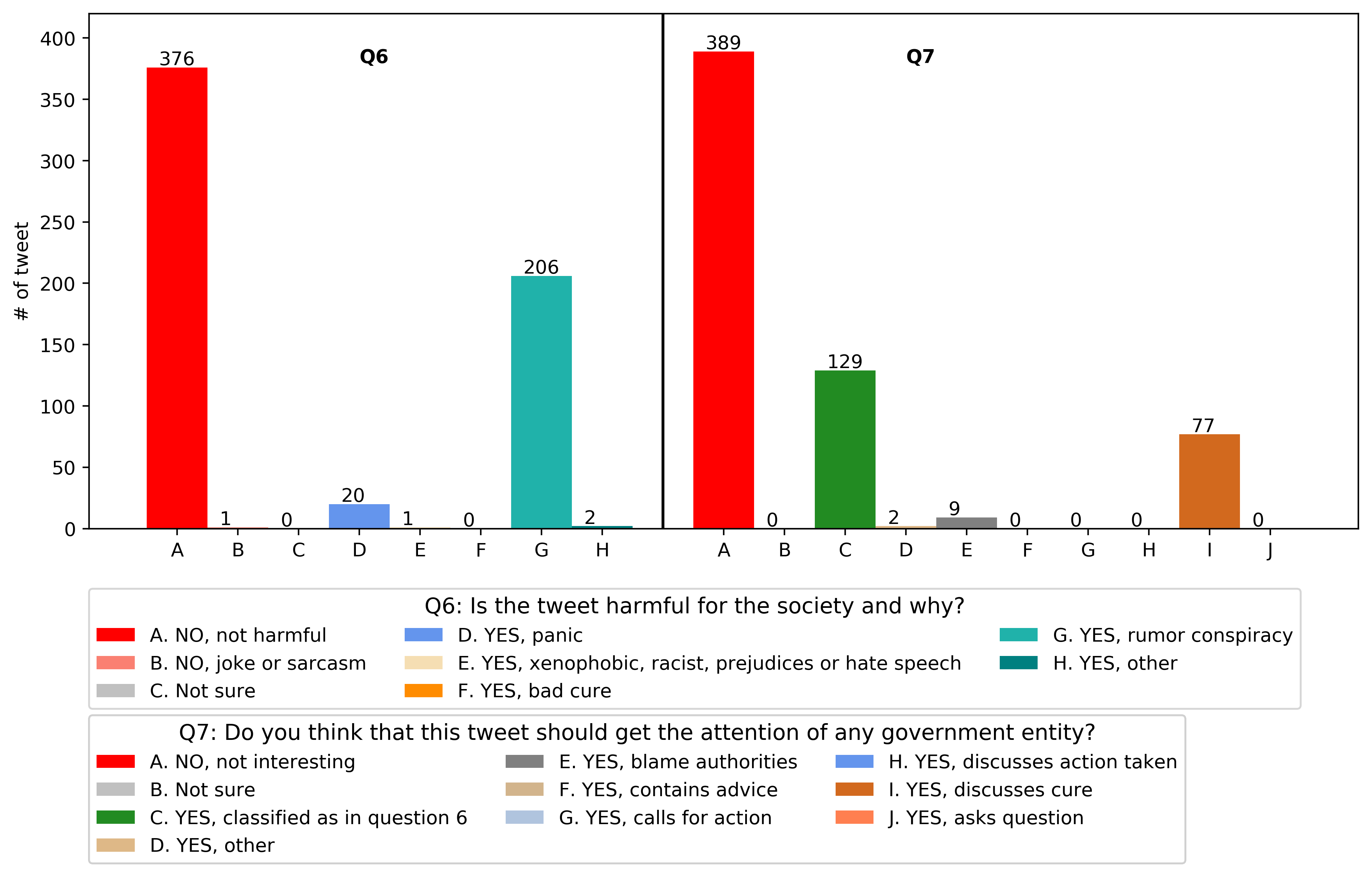}
        \caption{Questions Q6-Q7.}
        \label{fig:data_dist_group2_arabic}
    \end{subfigure}
    \caption{Statistics about the distribution of the \textbf{Arabic tweets} from February 2020 till March 2021.}
    \label{fig:data_dist_arabic_disinfo}
\end{figure*}

\begin{figure*}[tbh!]
\centering
    \begin{subfigure}[b]{0.75\textwidth}
        \includegraphics[width=\textwidth]{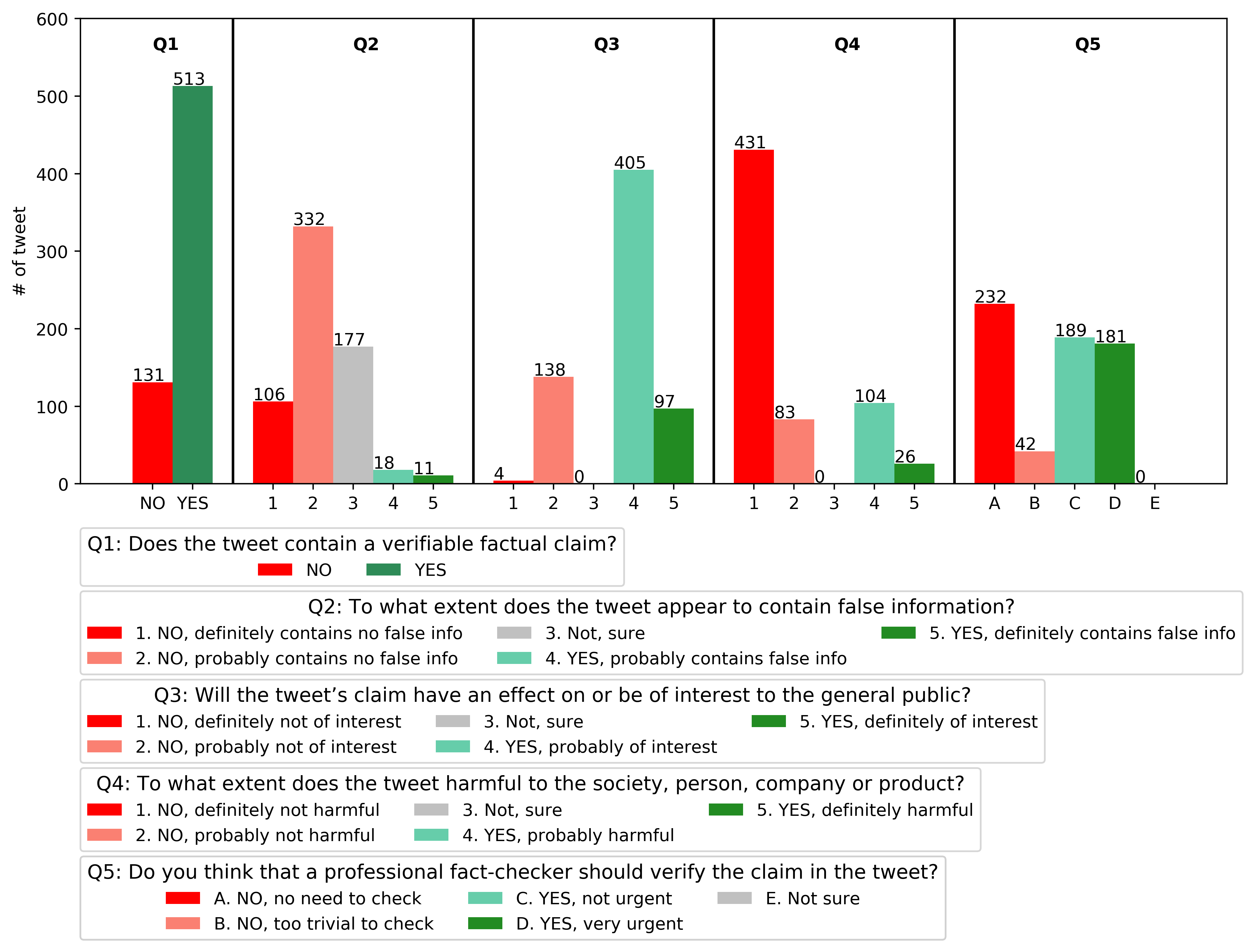}
        \caption{Questions Q1-Q5.}
        \label{fig:data_dist_group1_english}
    \end{subfigure}
    \hfill
    \begin{subfigure}[b]{0.75\textwidth}    
        \includegraphics[width=\textwidth]{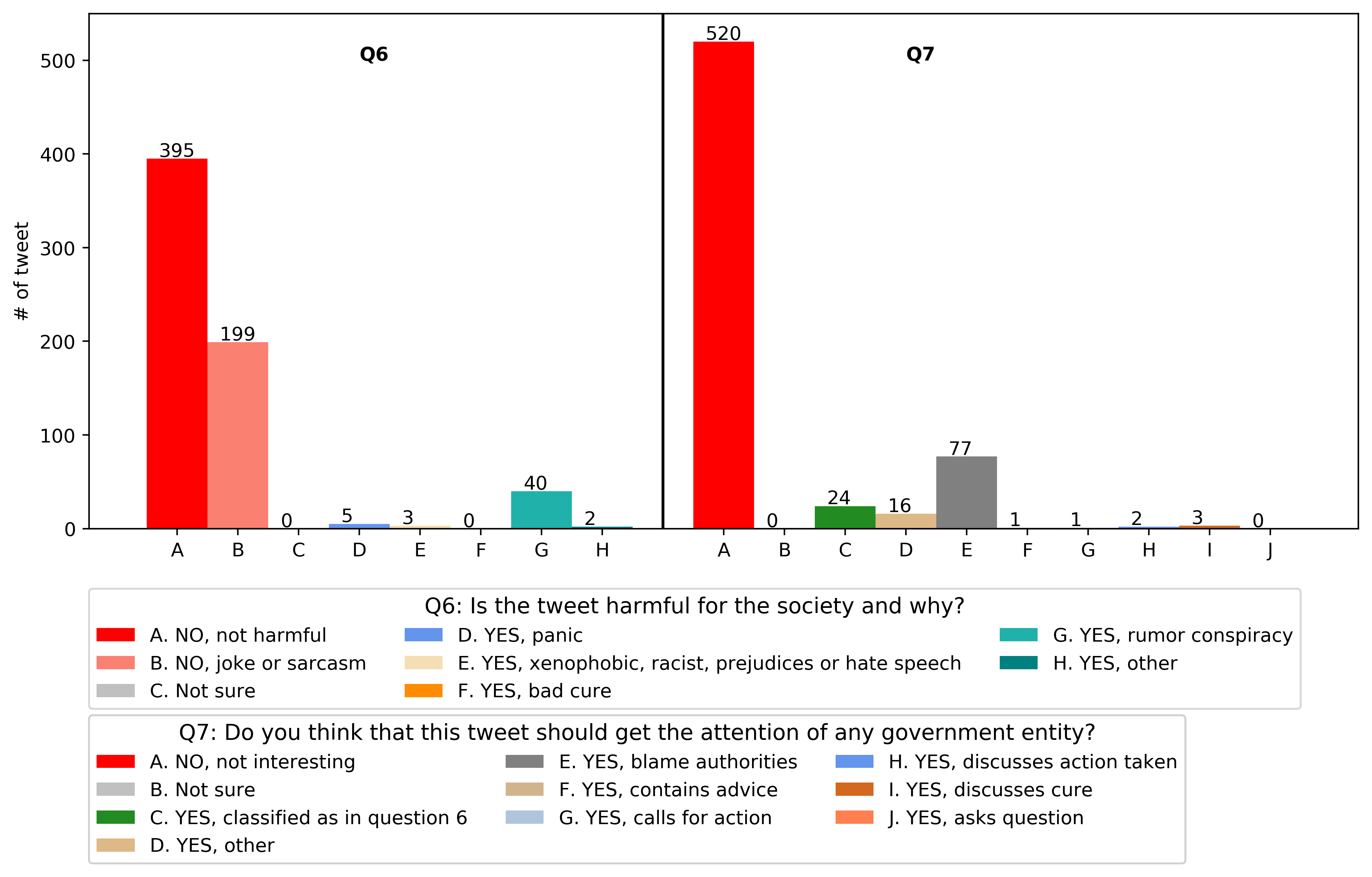}
        \caption{Questions Q6-Q7.}
        \label{fig:data_dist_group2_english}
    \end{subfigure}
    \caption{Statistics about the distribution of the \textbf{English tweets} collected from February 2020 till August 2020.}
    \label{fig:data_dist_english_disinfo}
\end{figure*}

\begin{figure*}[tbh!] 
\centering
    \begin{subfigure}[b]{0.75\textwidth}
        \includegraphics[width=\textwidth]{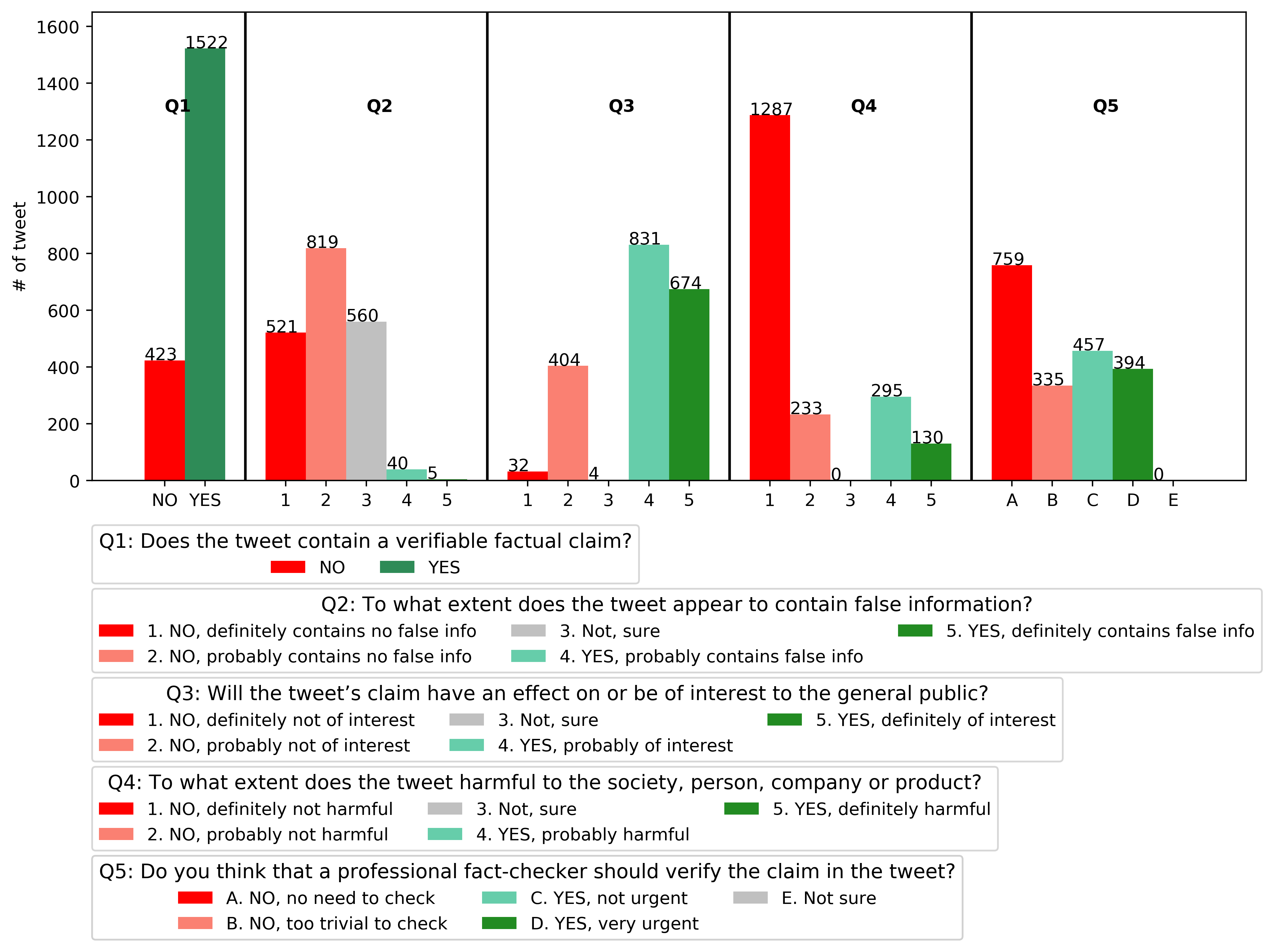}
        \caption{Questions Q1-Q5.}
        \label{fig:data_dist_group1_english_nov_20_jan_21}
    \end{subfigure}
    \hfill
    \begin{subfigure}[b]{0.75\textwidth}    
        \includegraphics[width=\textwidth]{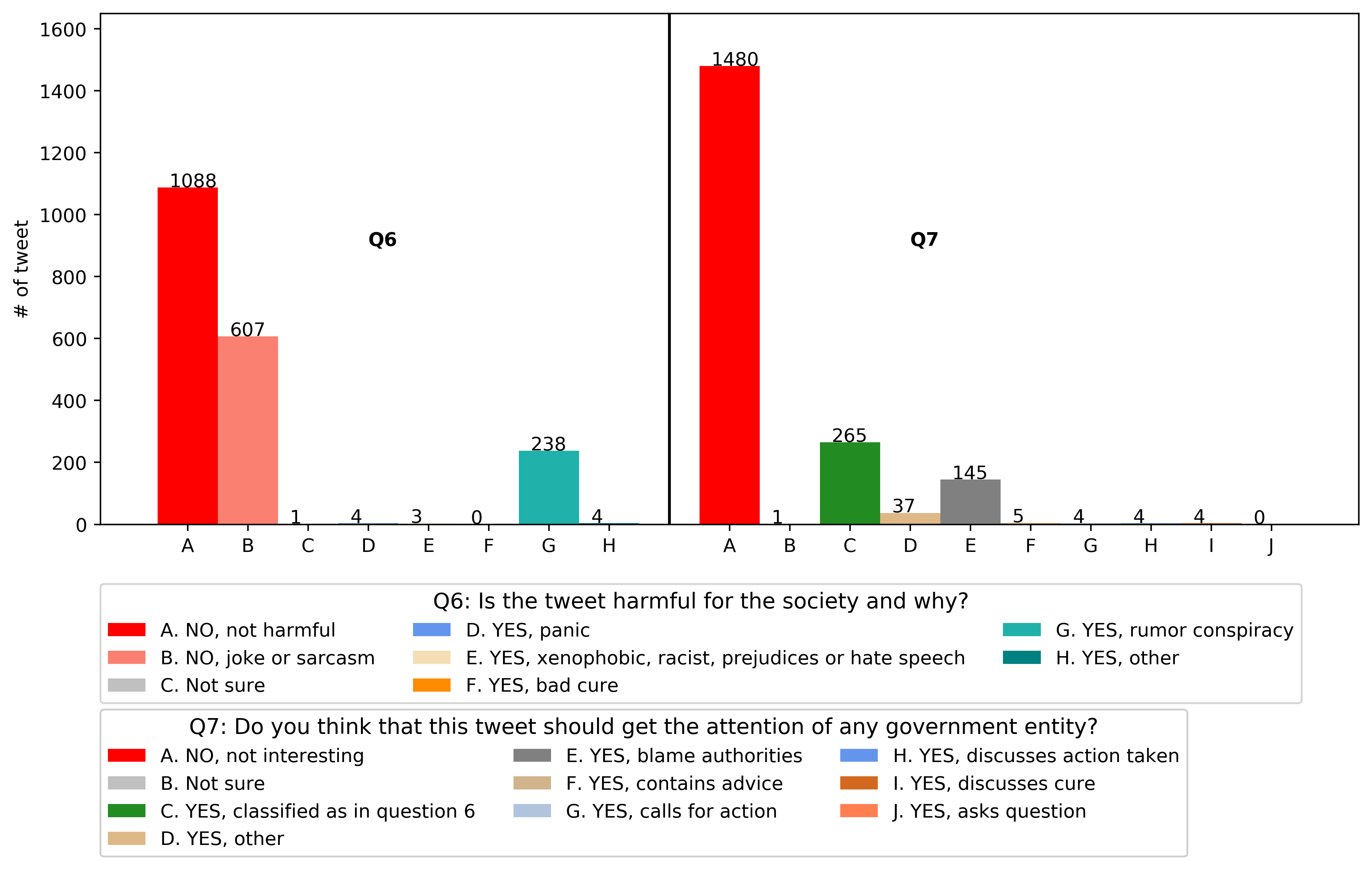}
        \caption{Questions Q6-Q7.}
        \label{fig:data_dist_group2_english_nov_20_jan_21}
    \end{subfigure}
    \caption{Statistics about the distribution of the \textbf{English tweets} collected from November 2020 till January 2021.}
    \label{fig:data_dist_english_disinfo_nov_20_jan_21}
\end{figure*}

\subsection{Framing}
\label{ssec:media_framing_analysis}

We used the \emph{Tanbih Framing Bias Detection system} \cite{zhang-etal-2019-tanbih}, trained on the Media Frames Corpus (11k training news articles) by fine-tuning BERT to detect topic-agnostic media frames, achieving accuracy of 66.7\% on the test set (1,138 news articles). It can predict the following 15 frames: (\emph{i})~Economy, (\emph{ii})~Capacity and resources, (\emph{iii})~Morality, (\emph{iv})~Fairness and equality, (\emph{v})~Legality, constitutionality and jurisprudence, (\emph{vi})~Policy prescription and evaluation, (\emph{vii})~Crime and punishment, (\emph{viii})~Security and defense, (\emph{ix})~Health and safety, (\emph{x})~Quality of life, (\emph{xi})~Cultural identity, (\emph{xii})~Public opinion, (\emph{xiii})~Politics, (\emph{xiv})~External regulation and reputation, and (\emph{xv})~Other.

\section{Results and Discussion}
\label{sec:results_discussions}

\subsection{Disinformation Analysis}

\paragraph{Arabic:} Figure \ref{fig:data_dist_arabic_disinfo} shows the distribution for the questions for Arabic. We can see that (\emph{i})~most tweets contain a verifiable factual claim, (\emph{ii})~about half of the tweets contain false information, (\emph{iii})~most tweets are of general interest to the public, (\emph{iv})~about half of the tweets are harmful to the society, a person, a company, or a product (Question 6), (\emph{v})~many tweets are worth fact-checking, (\emph{vi})~most tweets are not harmful to the society, and many spread rumors, and (\emph{vii})~some tweets discuss possible cure, and very few spread panic. 

\paragraph{English:} Figure \ref{fig:data_dist_english_disinfo} shows the distribution for the English tweets from February till August 2020. We can see that most tweets contain a verifiable factual claim, contain no false information, are of general interest to the public, are not harmful, and are worth fact-checking. Moreover, many tweets contain jokes, some contain rumors, and some blame the authorities. 

We also analyzed the English tweets from November 2020 till January 2021. The results are shown in Figure~\ref{fig:data_dist_english_disinfo_nov_20_jan_21}, and follow a very similar trend.

\paragraph{Summary:} Arabic tweets contain relatively more false information and rumors, some discuss possible cure, and very rarely spread panic. English tweets contain mostly factual statements, many make jokes, and rarely spread rumors. 

\begin{figure}[h]
\centering
    \begin{subfigure}[b]{0.3\textwidth}
        \includegraphics[width=\textwidth]{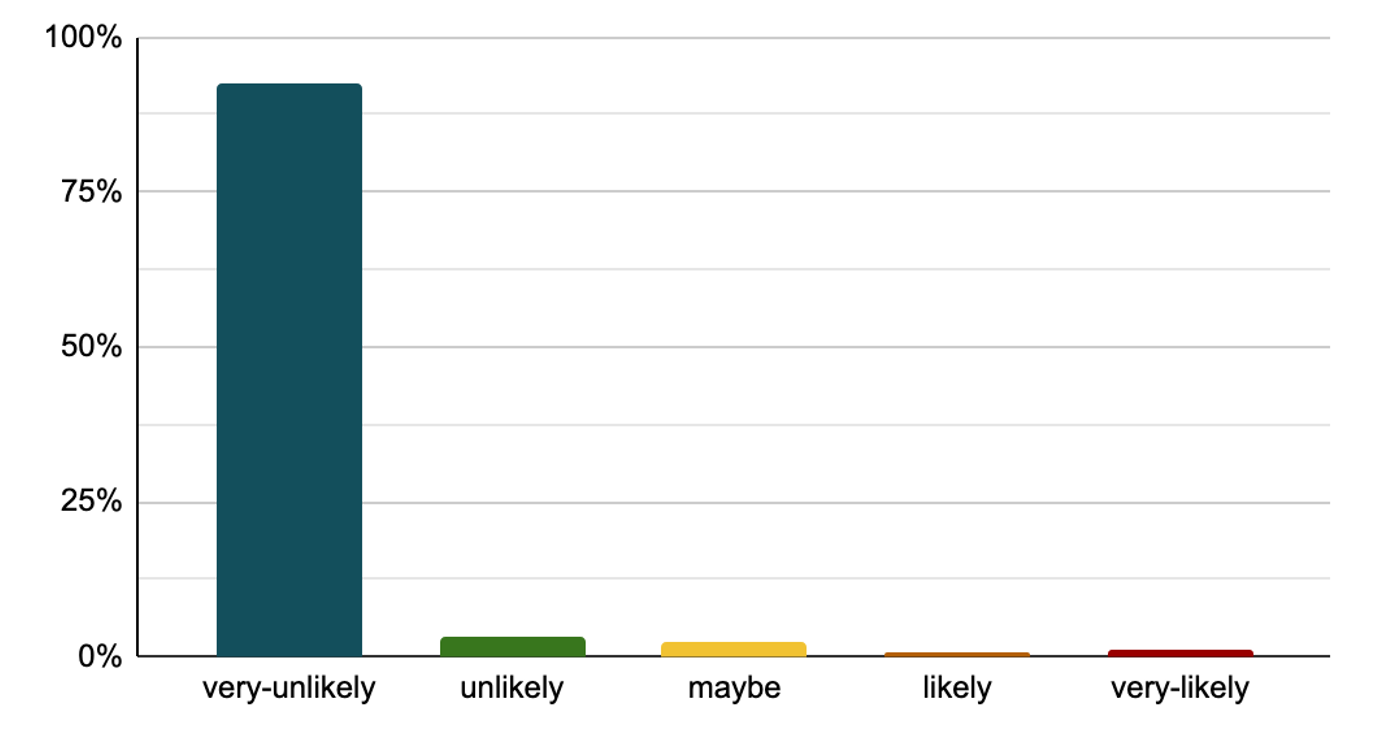}
        \caption{Arabic: February till August 2020}
        \label{fig:arabic_propaganda}
    \end{subfigure}
    \begin{subfigure}[b]{0.3\textwidth}    
        \includegraphics[width=\textwidth]{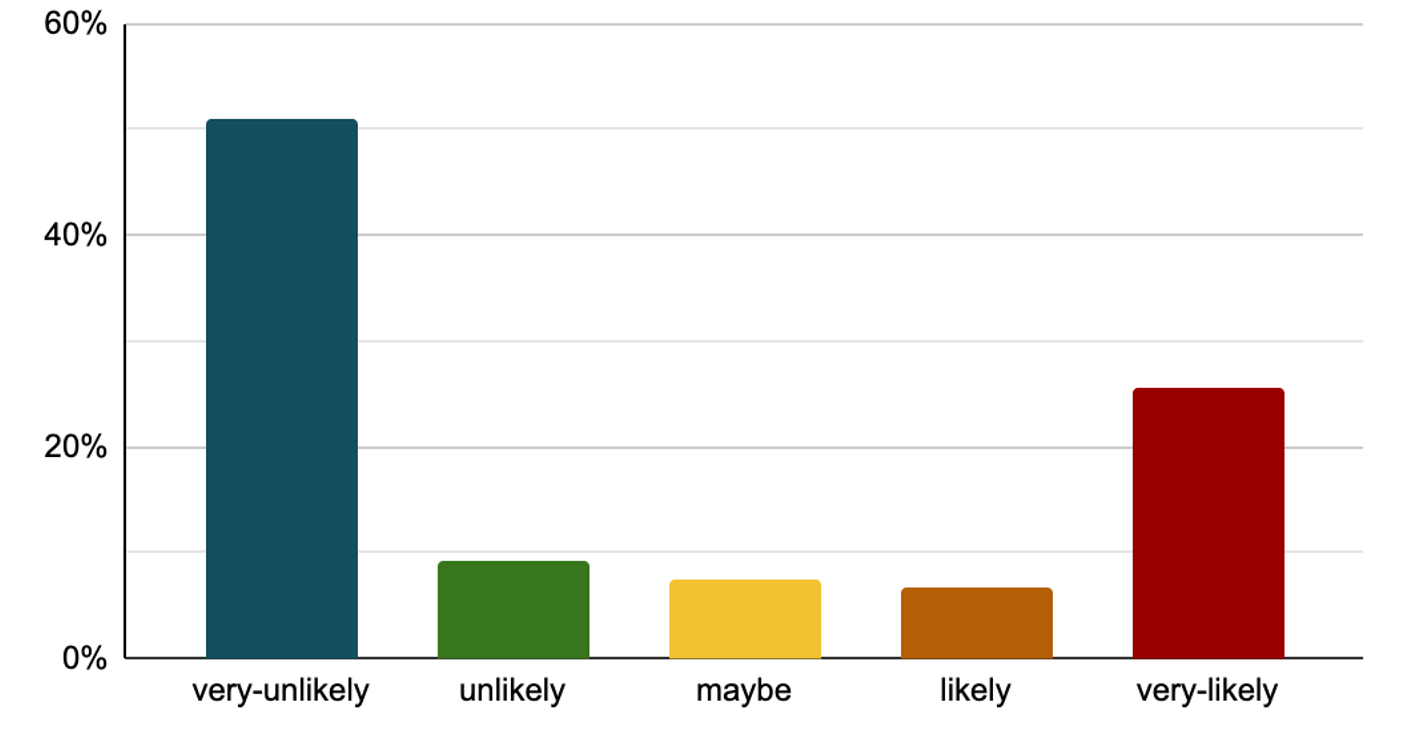}
        \caption{English: February till August 2020}
        \label{fig:english_propaganda}    
    \end{subfigure} 
    \begin{subfigure}[b]{0.3\textwidth}    
        \includegraphics[width=\textwidth]{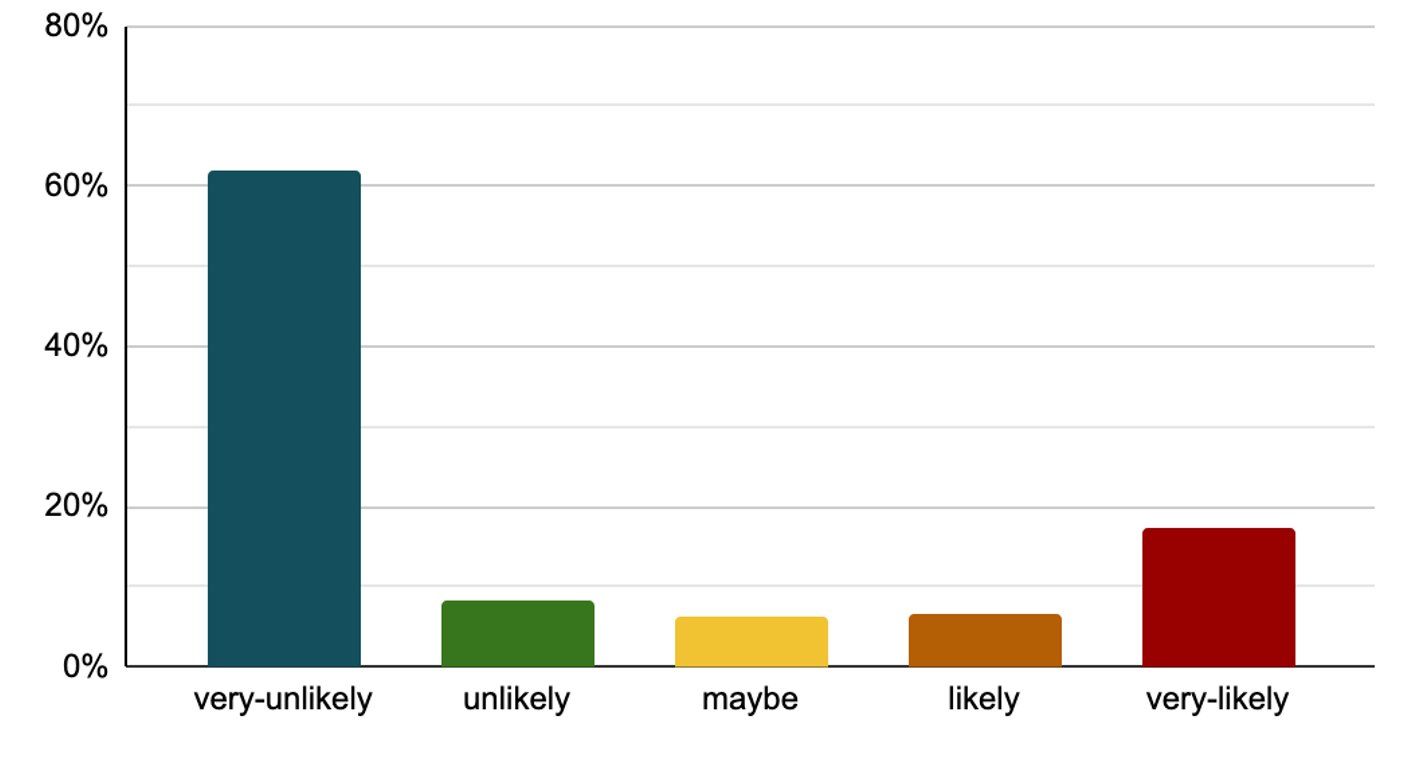}
        \caption{English: November 2020 till January 2021}
        \label{fig:english_propaganda_nov_20_jan_21}    
    \end{subfigure}
    \caption{Analysis: use of propaganda.}
    \label{fig:propanganda__ar_en_20}
\end{figure}

\subsection{Propaganda Analysis}

Figure \ref{fig:propanganda__ar_en_20} shows the propaganda analysis in Arabic vs. English tweets. We can see that Arabic propagandistic tweets are extremely rare, while for English they about 33\% of all tweets. 

We also analyzed English tweets collected from November 2020 till January 2021, to cover tweets about COVID-19 \emph{vaccines}, and we found that there were fewer propagandistic tweets: about 25\%.

\paragraph{Fine-Grained Propaganda Analysis}

\begin{figure}[t]
\centering
    \begin{subfigure}[b]{0.48\textwidth}
        \includegraphics[width=\textwidth]{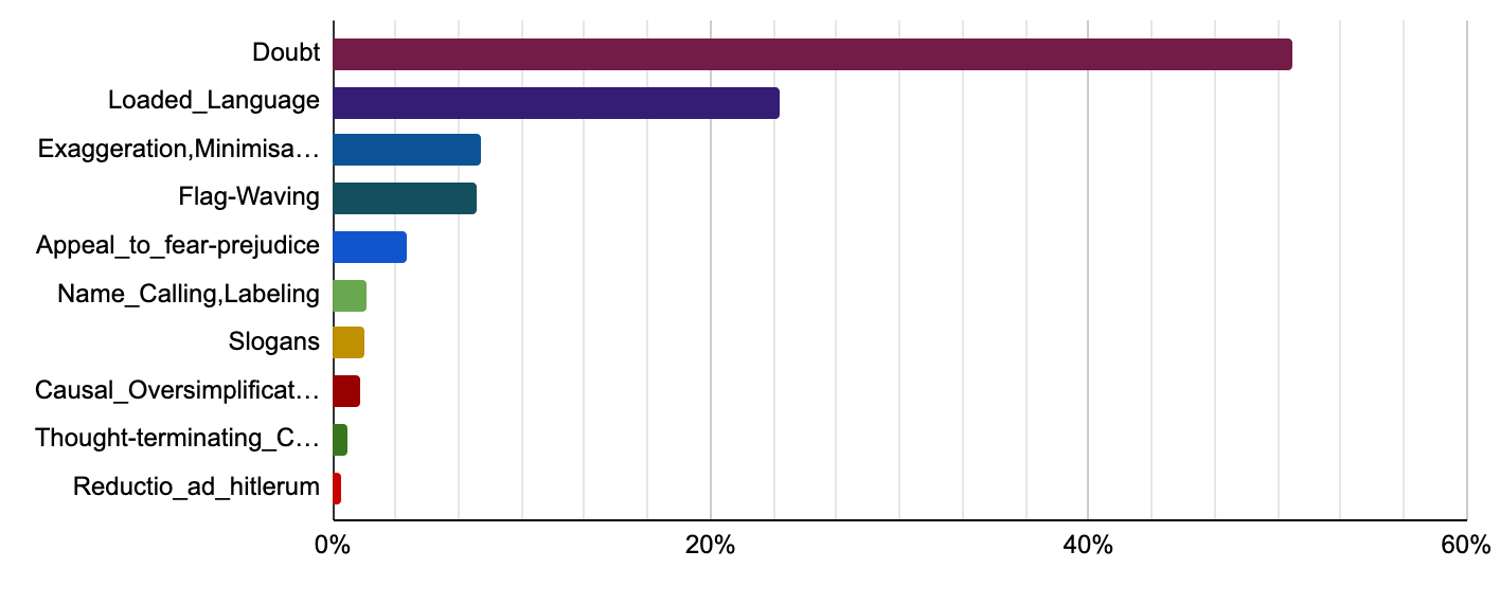}
        \caption{Arabic: February till August 2020}
        \label{fig:arabic_prop_tech}
    \end{subfigure}
    \begin{subfigure}[b]{0.48\textwidth}    
        \includegraphics[width=\textwidth]{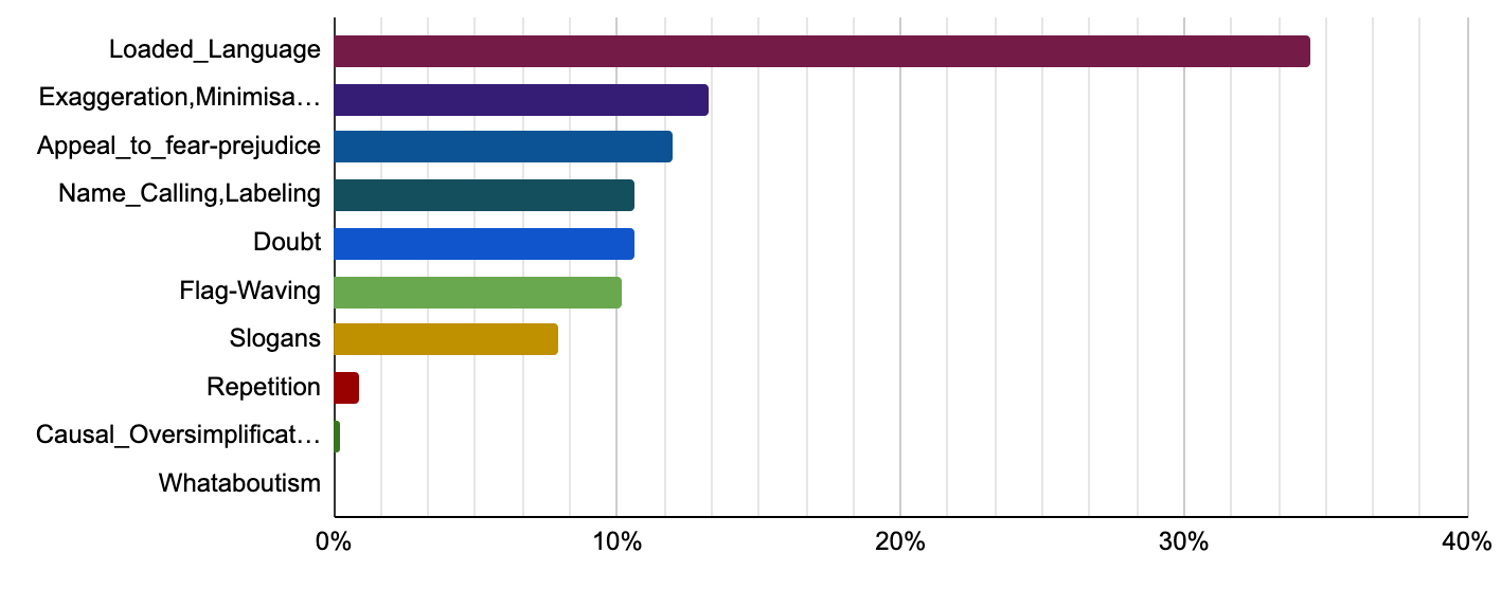}
        \caption{English: February till August 2020}
        \label{fig:english_prop_tech}    
    \end{subfigure} 
    \begin{subfigure}[b]{0.48\textwidth}    
        \includegraphics[width=\textwidth]{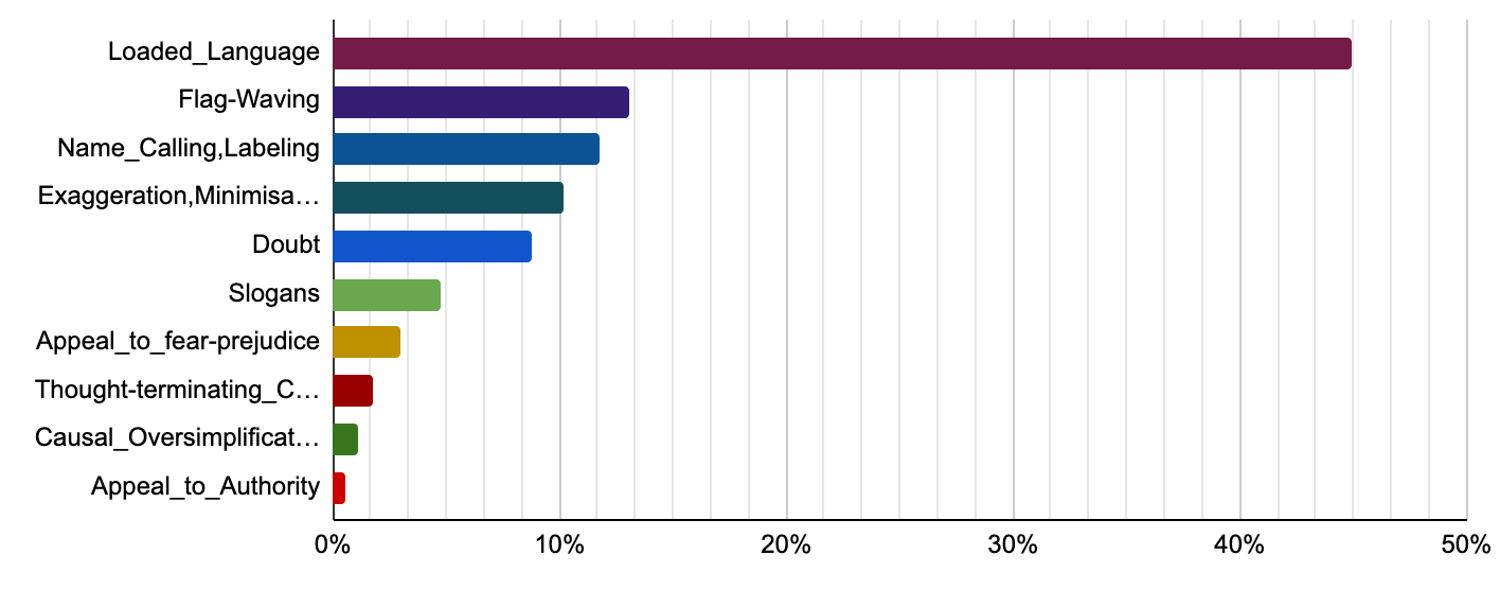}
        \caption{English: November 2020 till January 2021.}
        \label{fig:english_prop_tech_nov_20_jan_21}    
    \end{subfigure}
    \caption{Analysis: propaganda techniques.}
    \label{fig:propanganda_fine_grained}
\end{figure}

Next, we aimed to detect the specific propaganda techniques used in the tweets. Figure~\ref{fig:propanganda_fine_grained} shows the top propaganda techniques for Arabic and English. 

We can see that, for Arabic, 50\% of the tweets express \emph{doubt}, and 20\% use \emph{loaded language}. 

For English, we see a different distribution: about 33\% of the tweets use \emph{loaded language}, while each of the following techniques appears in about 10\% of the tweets: \emph{exaggeration}, \emph{fear}, \emph{name-calling}, \emph{doubt}, and \emph{flag-waving}.

Yet another trend is observed for English tweets collected from November 2020 till January 2021 (discussing vaccines): 50\% of the tweets use \emph{loaded language}, and each of the following four techniques appears in about 10\% of the tweets: \emph{flag-waving}, \emph{name-calling}, and \emph{exaggeration}.

\subsection{Framing}

\begin{figure}[h!]
\centering
    \begin{subfigure}[b]{0.48\textwidth}
        \includegraphics[width=\textwidth]{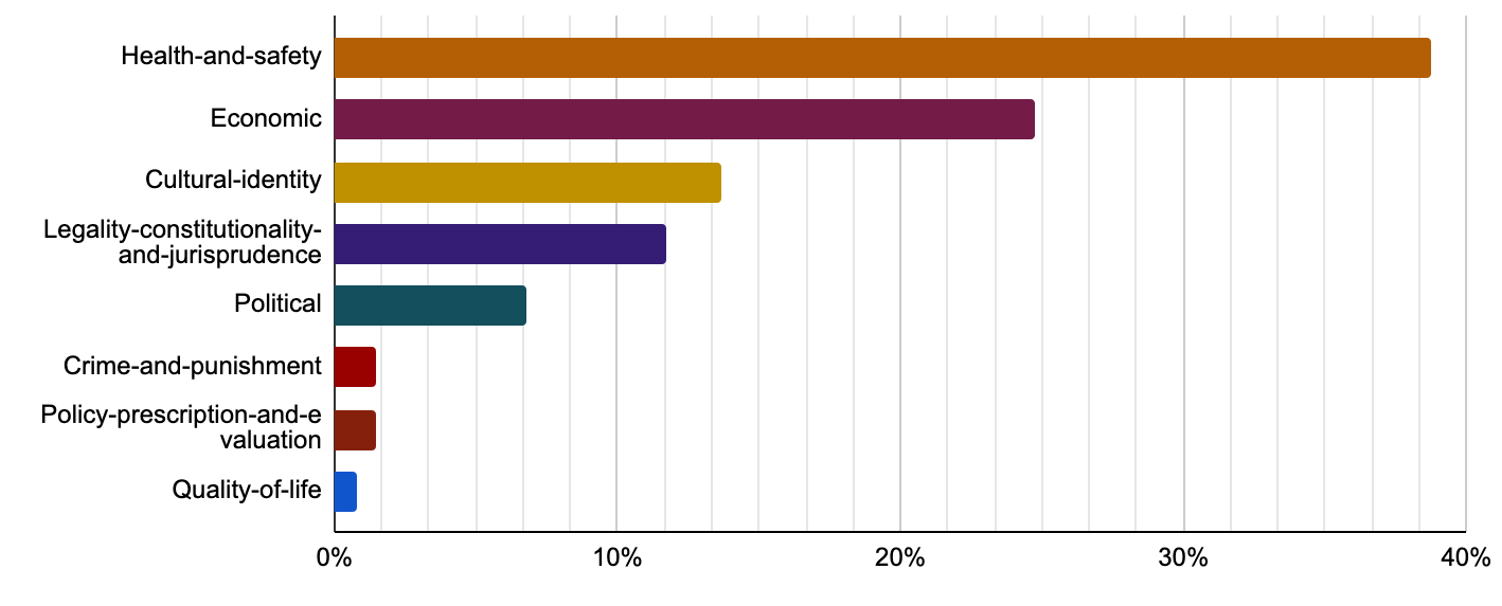}
        \caption{Arabic: February till August 2020.}
        \label{fig:arabic_framing}
    \end{subfigure}
    \begin{subfigure}[b]{0.48\textwidth}    
        \includegraphics[width=\textwidth]{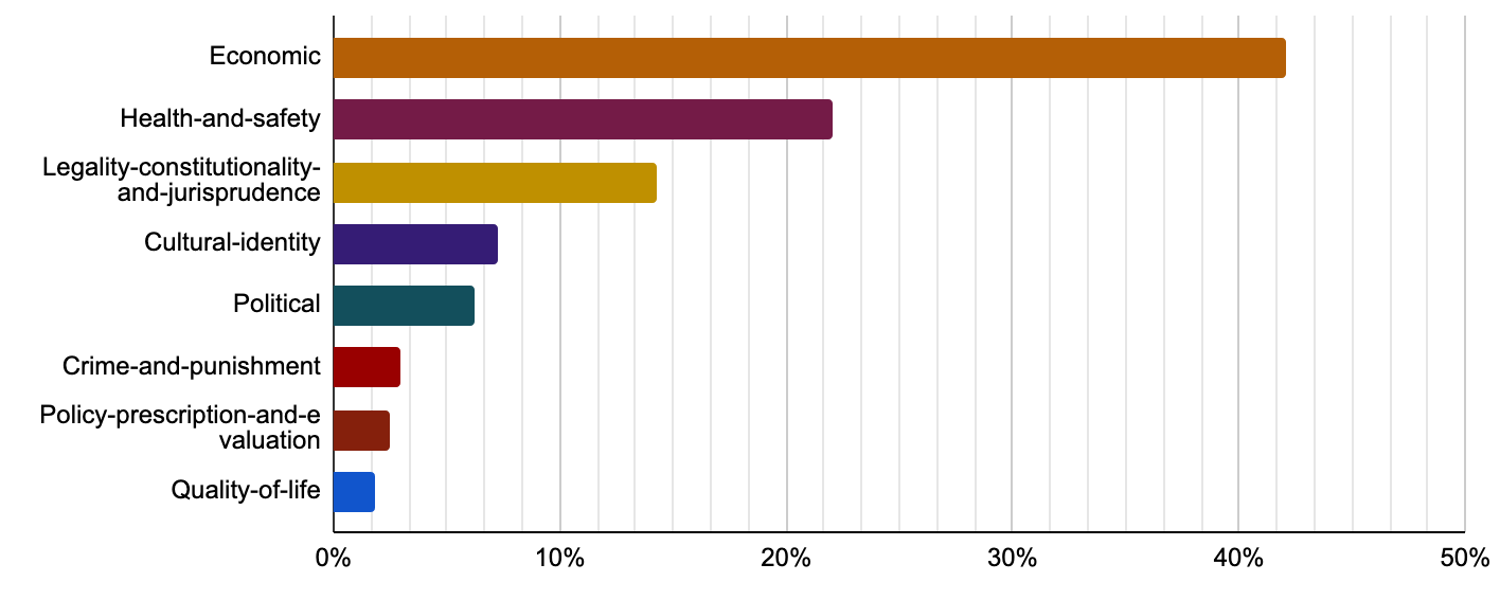}
        \caption{English: February till August 2020.}
        \label{fig:english_framing_feb_aug_20}    
    \end{subfigure} 
    \begin{subfigure}[b]{0.48\textwidth}    
        \includegraphics[width=\textwidth]{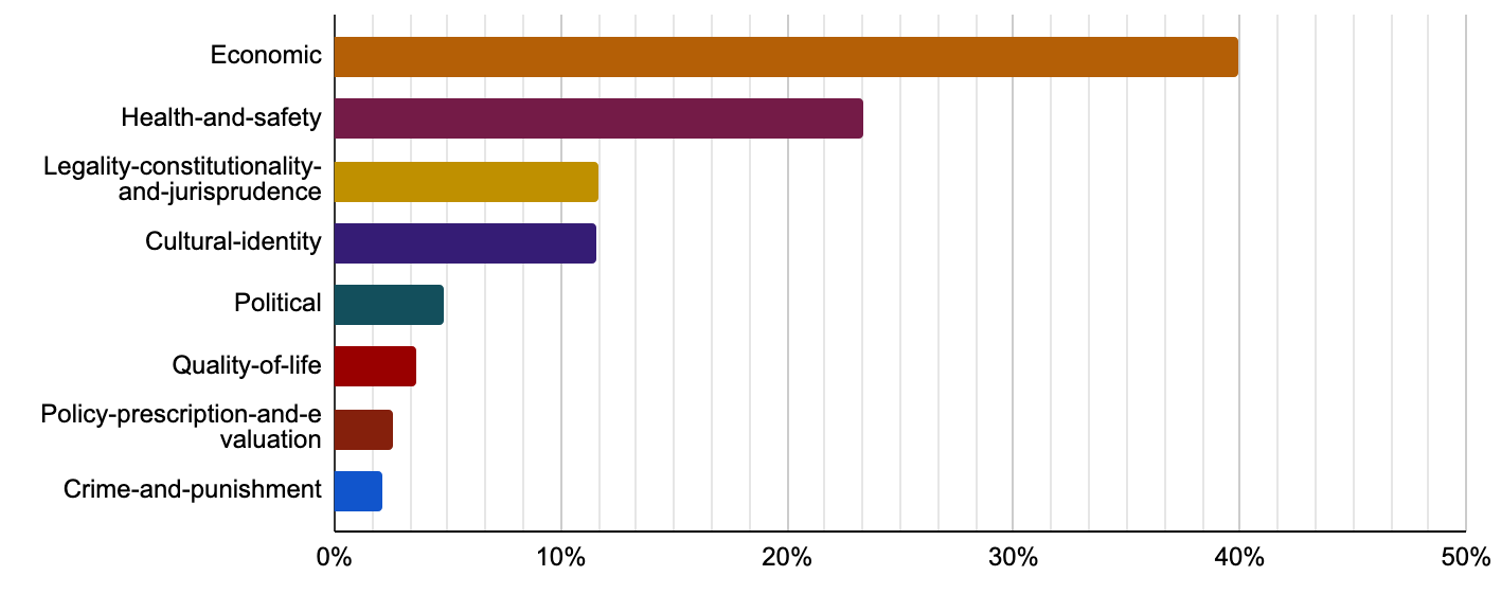}
        \caption{English: November 2020 till January 2021.}
        \label{fig:english_framing_nov_20_jan_21}    
    \end{subfigure}%
    \caption{Analysis: framing.}
    \label{fig:framing}
\end{figure}

Finally, we performed analysis in terms of framing, which reflects the perspective taken in the COVID-19 related tweets we analyzed. The results are shown on Figure~\ref{fig:framing}.

We can see that in the  Arabic tweets \emph{health and safety} is the dominant frame, with \emph{economy} coming second, and \emph{cultural identity} being third. 

For English, in both studied time periods, \emph{economy} is the primary frame, and \emph{health and safety} comes second. 

We speculate that the difference in framing between Arabic and English tweets reflects the perspective of Qatari locals (who tweet primarily in Arabic) vs. that of expats (who tweet primarily in English). Thus, it is to be expected that the former are concerned primarily with health aspects (e.g.,~COVID-19 vaccination, social distancing, and other measures to keep one safe during the pandemic), while the latter worry more about the economic consequences of the pandemic (and respectively, about the security of their jobs).

\section{Conclusion and Future Work}
\label{sec:conclutions}

We have presented our analysis of COVID-19 tweets in Arabic and English aiming to help in the fight against the global infodemic, which emerged as a result of the COVID-19 pandemic. In particular, we collected tweets in different time frames starting from February 2020 till January 2021, and we analyzed them using different aspects of disinformation, propaganda, and framing. We believe that such analysis should help in better understanding the trends over time and across languages.

Many interesting directions could be pursued in future work. For example, the analysis could be applied to other languages; in fact, we already did a related study for Bulgarian \cite{RANLP2021:COVID19:Bulgarian}. Moreover, while here we focused on tweets, the approach is applicable to other social media platforms such as Facebook and WhatsApp.

\section*{Acknowledgments}

This research is part of the Tanbih mega-project,\footnote{\url{http://tanbih.qcri.org}} developed at the Qatar Computing Research Institute, HBKU, which aims to limit the impact of ``fake news'', propaganda, and media bias by making users aware of what they are reading, thus promoting media literacy and critical thinking.

\bibliographystyle{acl_natbib}
\bibliography{bib/all_bib}

\end{document}